\documentclass[10pt,twocolumn,letterpaper]{article}

\usepackage{cvpr}              % To produce the CAMERA-READY version
\definecolor{cvprblue}{rgb}{0.21,0.49,0.74}
\usepackage[pagebackref,breaklinks,colorlinks,allcolors=cvprblue]{hyperref}

\def\paperID{*****} % *** Enter the Paper ID here
\def\confName{CVPR}
\def\confYear{2026}

\title{LOBSTgER-enhance: an underwater image enhancement pipeline}

\author{Andreas Mentzelopoulos, Keith Ellenbogen\\
Massachusetts Institute of Technology\\
77 Massachusetts Ave, Cambridge, MA, 02139\\
{\tt\small ament@mit.edu, keithe@mit.edu}
}

\begin{document}
\maketitle
\begin{abstract}
Underwater photography presents significant inherent challenges including reduced contrast, spatial blur, and wavelength-dependent color distortions. These effects can obscure the vibrancy of marine life and awareness photographers in particular are often challenged with heavy post-processing pipelines to correct for these distortions.

We develop an image-to-image pipeline that learns to reverse underwater degradations by introducing a synthetic corruption pipeline and learning to reverse its effects with diffusion-based generation. Training and evaluation are performed on a small high-quality dataset of awareness photography images by Keith Ellenbogen. The proposed methodology achieves high perceptual consitency and strong generalization in synthesizing 512x768 images using a model of $\sim$11M parameters after training from scratch on $\sim$2.5k images.

\end{abstract}    
\section*{Keywords}

Latent diffusion, image-to-image, generative-AI, underwater image restoration, LOBSTgER.

\vspace{0.5em}\noindent
\textbf{Github-repo}: \href{https://github.com/mentzelopoulos/Latent_UnderWater_Diffusion}{LOBSTgER-enhance}

\section{Introduction}
\label{sec:intro}

\begin{figure}[t]
    \centering
    \includegraphics[width=\linewidth]{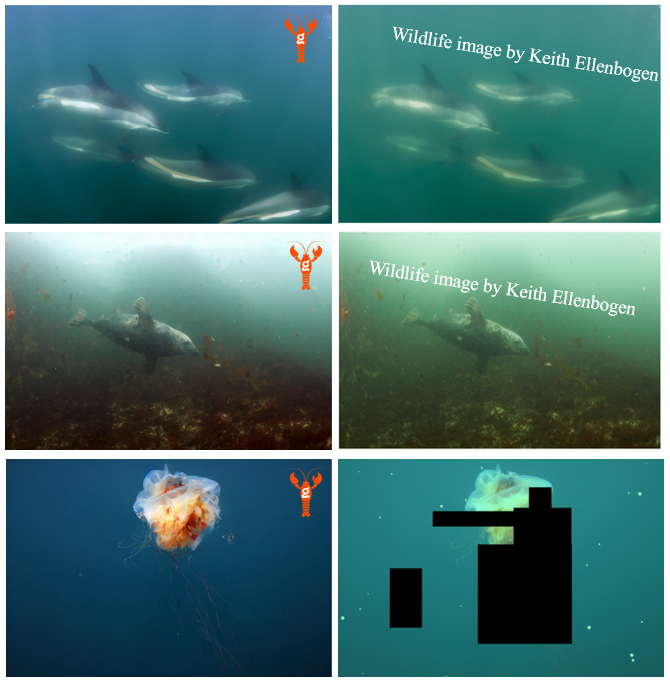}
    \caption{Inference samples for image enhancement and inpainting with LOBSTgER-enhance. Generated samples are shown on the left column while conditions are given in the right column. The model has never seen dolphins or seals during training. Sample dimensions are 512x768.}
    \label{fig:LOBSTgER_samples}
\end{figure}

Photography has long been a powerful medium for raising awareness about social and environmental issues, capturing compelling moments, and telling stories that resonate with audiences. Early conservation photographers, such as Carleton Watkins, showcased the grandeur of natural landscapes like Yosemite, influencing the creation of national parks and shaping public conservation efforts. Similarly, Jacques Cousteau and Harold 'Doc' Edgerton, expanded visual storytelling to the underwater realm using innovative lighting techniques, revealing the ocean’s beauty and making it accessible to the public. These historical examples highlight how technical innovation and artistic vision can combine to create enduring impact.

Generative AI has emerged as a transformative technology for visual media, enabling the creation of highly realistic images that can complement or coexist with traditional photography. With the increasing accessibility and realism of AI-generated imagery, there is an opportunity to harness these models for environmental advocacy, creating visually compelling content that educates and engages diverse audiences. Generative approaches allow for scalable, high-fidelity synthesis of imagery, offering new ways to amplify environmental narratives while maintaining artistic integrity.

LOBSTgER builds on this legacy by pairing high-quality underwater awareness photography with generative latent diffusion models to produce authentic, high-resolution synthetic imagery. This integration of photography and AI enables LOBSTgER to generate compelling visual stories that bridge real and synthetic imagery, advancing environmental awareness and advocacy in the digital age.

\subsection{Related Literature}

Generative models, particularly in the form of diffusion models and GANs, have significantly advanced the enhancement of underwater images. Underwater imaging faces numerous challenges due to light absorption, scattering, and distortion caused by the aquatic environment, leading to poor visibility and image degradation \cite{duarte2016dataset}. 

Recent research has leveraged generative models to address these issues. Studies have been conducted to restore natural coloration to underwater scenes using both GANs \cite{wang2025large} and Diffusion models\cite{shi2024cpdm}. Other studies \cite{zhao2025generative} have incorporated novel mechanisms such as multiscale and attention layers and demonstrate how these techniques allow for more effective image enhancement, focusing on restoring fine details and boosting contrast. These methods provide better preservation of texture and feature details, resulting in more visually accurate and scientifically useful images.

Other innovations including  \cite{bach2024underwater} and \cite{zhao2024learning} incorporate physics based modeling to simulate the complex light conditions found underwater. By integrating these physical priors into their model design, they deliver more realistic and natural-looking images. These advancements collectively represent a major leap in utilizing generative models for improving underwater imaging, particularly for scientific research and commercial applications like underwater photography and deep-sea exploration.

\begin{figure}[tb]
    \centering
    \includegraphics[width=\linewidth]{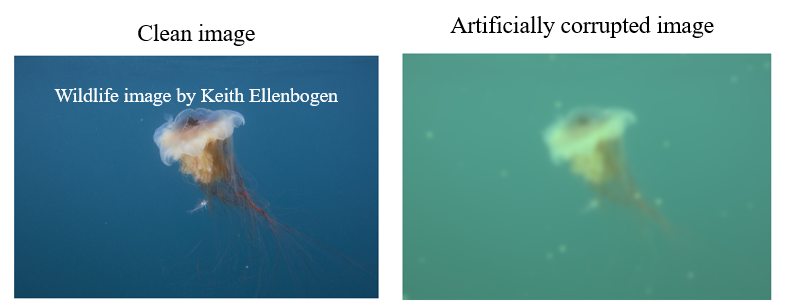}
    \caption{Artificial corruption process used to generate clean/corrupted pairs for conditional diffusion model supervised training. Left: Clean image by Keith Ellenbogen. Right: Corrupted image using the defined corruption pipeline: colors are distorted, bubbles are scattered throughout the image, hazing effect and gaussian blur are applied.}
    \label{fig:artificial_corruption}
\end{figure}

\section{Methodology \& Experiments}

In this work, we propose an novel pipeline for underwater image restoration via image-to-image diffusion. We first develop a synthetic corruption function that simulates underwater image degradation and then train a compact diffusion model to reverse these corruptions. We focus on images of size 512x768, with an aspect ratio of 2/3.

\subsection{Modeling Underwater Image Degradation}

To train our models, we generate paired datasets of clean and synthetically degraded underwater images. We design a corruption function that reproduces typical underwater distortions through multiple stages: (i) bubble artifacts, introducing randomly sized bright spots to simulate suspended particles or air bubbles; (ii) applying semi-transparent gray overlays to emulate light scattering from sediments and turbidity; (iii) motion blur, using randomized Gaussian kernels to model camera or water motion; (iv) color distortion, introducing random channel-wise shifts to mimic wavelength-dependent absorption; and (v) additional degradations, including vignetting, Gaussian noise, and gamma variations to capture lens and lighting effects. This pipeline produces realistic degraded images for supervised learning while maintaining alignment with real underwater conditions.

Figure~\ref{fig:artificial_corruption} illustrates a pair of clean and corrupted images using the artificial corruption process. On the left is the clean image, captured by Keith Ellenbogen, showcasing the natural colors, clarity, and fine-grained details of marine life under good visibility conditions. On the right is the corresponding corrupted version, produced using the defined corruption function. Color distortion is visible, with warmer hues such as red and orange significantly reduced, resulting in a bluish-green tint. Artificial bubble-like artifacts have been introduced to simulate particulate matter or air bubbles suspended in the water. A hazing layer has been applied to reduce contrast and replicate turbidity caused by sediments and plankton. Additionally, a mild Gaussian blur has been introduced to simulate the effects of light scattering and motion. These visual degradations collectively produce an approximation of challenging underwater imaging conditions, enabling the creation of paired training examples for supervised enhancement tasks. 

\subsection{Conditional Latent Diffusion}

\begin{figure}[tb]
    \centering
    \includegraphics[width=\linewidth]{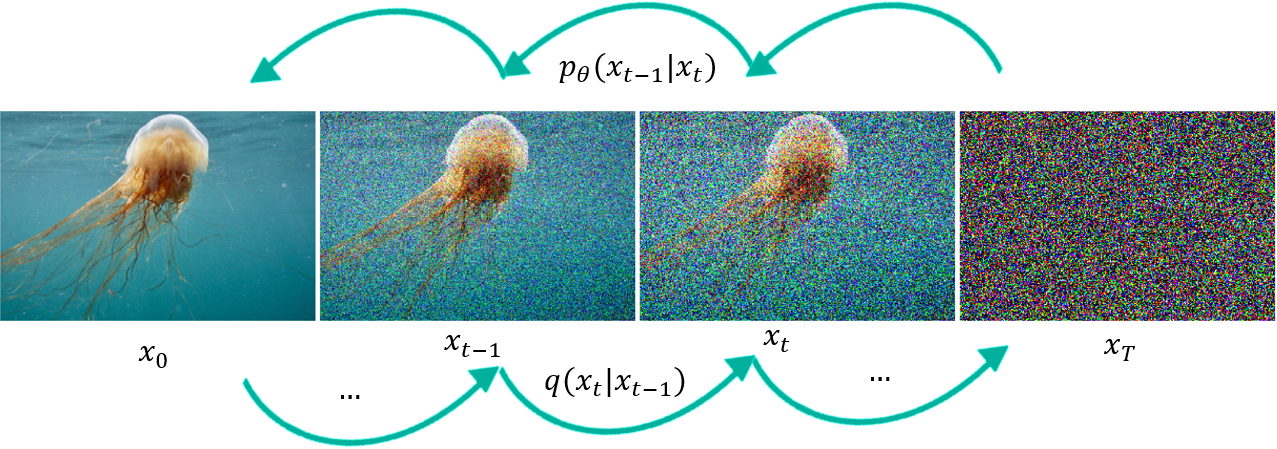}
    \caption{Illustration of forward and reverse diffusion process.}
    \label{fig:diffusion_intro}
\end{figure}

We build upon the Denoising Diffusion Probabilistic Model (DDPM) framework~\cite{ho2020denoising}, which defines a forward noising process and a learned reverse denoising process as illustrated in Figure \ref{fig:diffusion_intro}. Given an image \( \mathbf{x}_0 \), Gaussian noise is progressively added through
\begin{equation}
    q(\mathbf{x}_t|\mathbf{x}_{t-1}) = \mathcal{N}\!\left(\sqrt{1-\beta_t}\,\mathbf{x}_{t-1},\,\beta_t\mathbf{I}\right),
\end{equation}
where the cumulative product \(\bar{\alpha}_t = \prod_{s=1}^{t}(1-\beta_s)\) yields the closed-form
\begin{equation}
    q(\mathbf{x}_t|\mathbf{x}_0) = \mathcal{N}\!\left(\sqrt{\bar{\alpha}_t}\,\mathbf{x}_0,\,(1-\bar{\alpha}_t)\mathbf{I}\right).
\end{equation}
A neural network \(\epsilon_\theta(\mathbf{x}_t, t)\) is trained to predict the added noise using the MSE loss
\begin{equation}
    \mathcal{L} = \mathbb{E}_{\mathbf{x}_0,\epsilon,t}\!\left[\|\epsilon - \epsilon_\theta(\mathbf{x}_t,t)\|^2\right].
\end{equation}

\begin{figure}[tb]
    \centering
    \includegraphics[width=\linewidth]{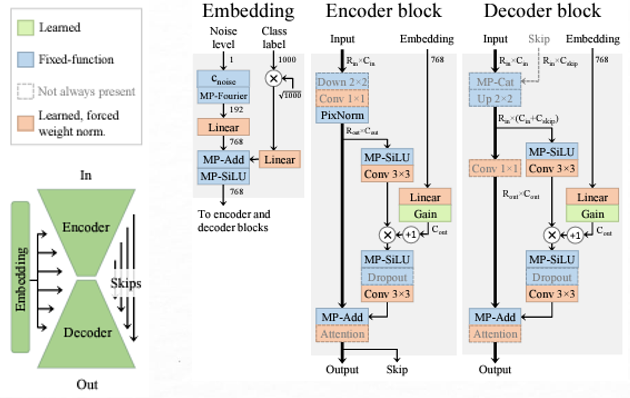}
    \caption{U-Net architecture adapted from~\cite{karras2024analyzing}.}
    \label{fig:karras_architecture}
\end{figure}

To improve training stability, we adopt the cosine variance schedule from~\cite{nichol2021improved}, which provides smoother corruption and better signal preservation than the linear schedule. Furthermore, we employ the velocity parameterization (\(v\)-prediction) popularized by~\cite{karras2022elucidating}, where
\begin{equation}
    \mathbf{v}_t = \sqrt{\bar{\alpha}_t}\,\epsilon - \sqrt{1-\bar{\alpha}_t}\,\mathbf{x}_0,
\end{equation}
and the model minimizes
\begin{equation}
    \mathcal{L}_\text{v-pred} = \mathbb{E}\!\left[\|\hat{\mathbf{v}}_\theta(\mathbf{x}_t,t) - \mathbf{v}_t\|^2\right].
\end{equation}

For conditional generation, we modify the reverse process to incorporate conditioning variables \(y\)
\begin{equation}
    p_\theta(\mathbf{x}_{t-1}|\mathbf{x}_t, y) = \mathcal{N}\!\big(\mu_\theta(\mathbf{x}_t,t,y),\,\Sigma_\theta(\mathbf{x}_t,t,y)\big),
\end{equation}
where conditioning is implemented via concatenation.

Finally, to reduce computational cost, diffusion operates in the latent space of a pretrained variational autoencoder (VAE) by \cite{rombach2022high}. The VAE encoder compresses high-resolution images into a lower-dimensional latent representation, on which diffusion is trained; the decoder subsequently reconstructs images from the refined latent samples.

In addition, following the approach of \cite{daras2025ambient}, we adopt a quality-weighted sampling strategy during diffusion training. Specifically, only the top $10\%$ of images, as ranked by Keith Ellenbogen (the photographer), were used across all diffusion timesteps, including those corresponding to low noise levels. The remaining $90\%$ of the dataset were included only at higher noise levels ($500 < t < 1000$). This design introduces a structural bias in the generative process toward the visual characteristics of the highest-quality samples.

\subsection{Diffusion model training}

\subsubsection{Neural Architecture}

Our diffusion backbone follows the U-Net architecture of Karras et~al.~\cite{karras2024analyzing}. Images are first encoded into latent representations using a pretrained VQ-GAN~\cite{esser2020taming}. Diffusion operates directly in this latent space.

The U-Net is intentionally lightweight, comprising two encoder and two decoder blocks (Fig.~\ref{fig:karras_architecture}) with roughly 11M trainable parameters. Latent inputs of four channels are progressively expanded through the encoder to 128 and 256 channels, enabling richer feature abstraction at deeper layers. The decoder mirrors this structure, performing symmetric upsampling to reconstruct the latent dimensionality.

\subsubsection{Training hyperparameters}

\begin{table}[tb]
  \caption{Training Hyperparameters}
  \label{tab:train_hyperparameters}
  \centering
  \begin{tabular}{@{}lc@{}}
    \toprule
    Hyperparameter & Value \\
    \midrule
    U-Net Architecture & Adapted from \cite{karras2024analyzing} \\
    Pre-trained VAE & from \cite{esser2020taming} \\
    \# train params & $\sim$ 11 M \\
    Diffusion timesteps & 1000 \\
    Loss & v-prediction \cite{karras2022elucidating}\\
    Variance schedule & Cosine \cite{nichol2021improved} \\
    Optimizer & AdamW \\
    Batch size & 64\\
    Learning rate & 1e-2 \\
    LR Scheduler & Custom flat-the-decay \\
    Train epochs & 15k \\
    Data sampling & Custom inspired by \cite{daras2025ambient} \\
    \bottomrule
  \end{tabular}
\end{table}

\begin{figure}[tb]
    \centering
    \includegraphics[width=0.75\linewidth]{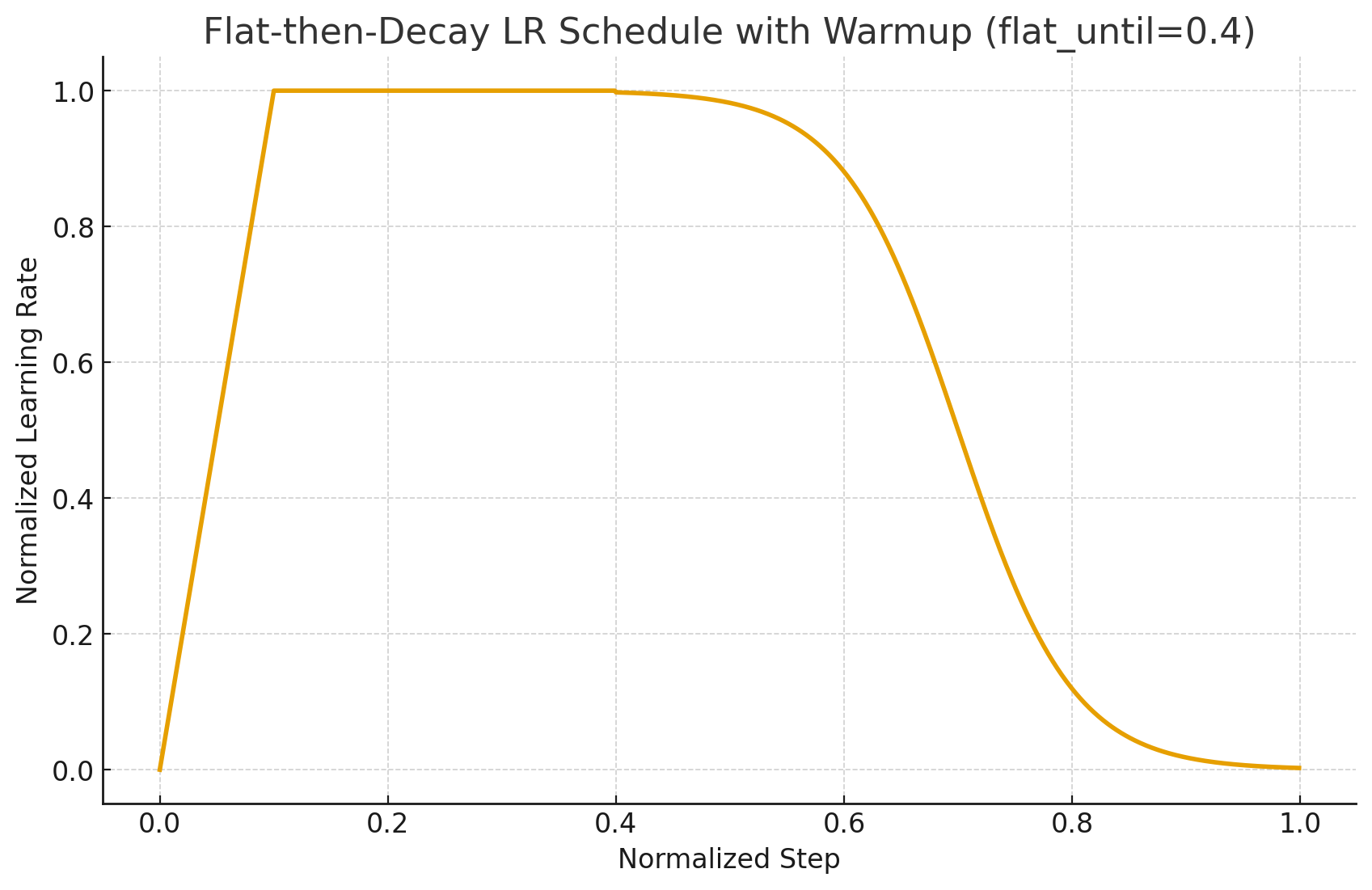}
    \caption{Normalized learning rate schedule.}
    \label{fig:lr_scheduler}
\end{figure}

\begin{figure}[tb]
    \centering
    \includegraphics[width=\linewidth]{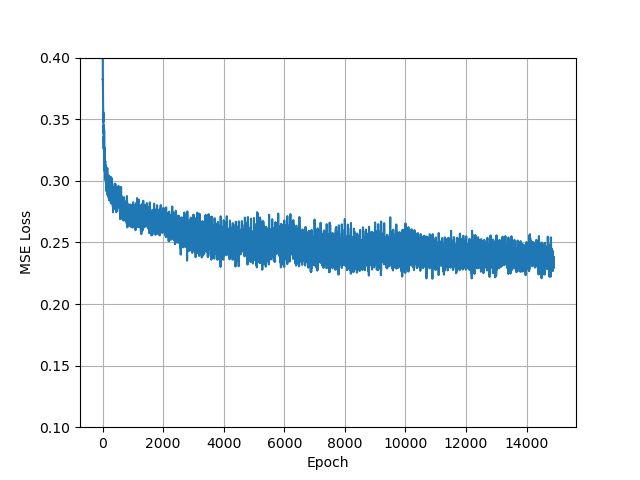}
    \caption{Loss function over 15k epochs of training.}
    \label{fig:Loss_diffusion}
\end{figure}

Model parameters were optimized using the AdamW optimizer with an aggressive learning rate of 1e-2. The learning rate followed a flat-then-decay schedule (shown in Figure \ref{fig:lr_scheduler}) with a brief warmup period, allowing for stable convergence in early iterations and gradual annealing during the second half of training. A batch size of 64 was used for all experiments. Training was done on an NVIDIA Tesla V100 GPU with 32~GB of memory using mixed-precision arithmetic to accelerate computation and reduce memory consumption.  The training loss over 15k epochs is shown in Figure \ref{fig:Loss_diffusion}. Convergence is achieved.

\subsection{Training loop}

\begin{figure}[b]
    \centering
    \includegraphics[width=\linewidth]{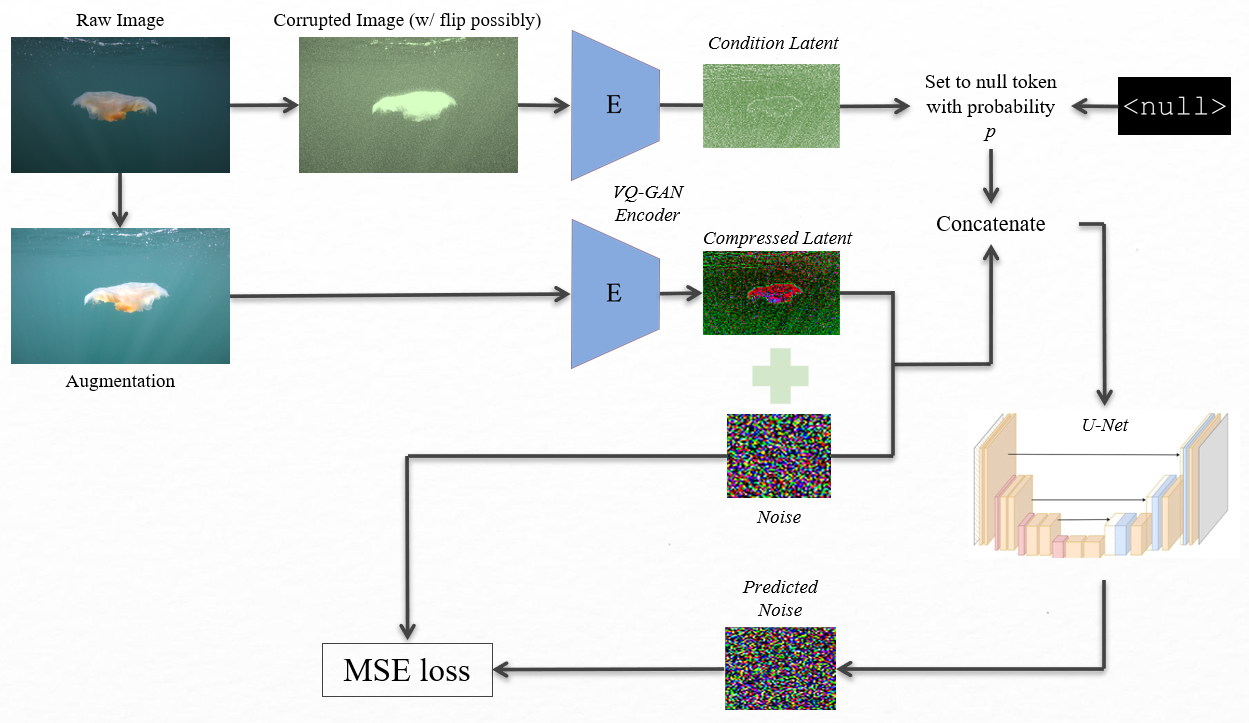}
    \caption{Visualization of the conditional model training loop. Raw images were first augmented and corrupted to obtain augmented-corrupted pairs. The pairs were passed through the encoder to obtain their latent representations. The U-Net was tasked to learn the noise added to the (clean) augmentation given the noisy latent and the condition latent, combined through concatenation. In order to enable classifier-free guidance, the condition latent was set to a (learnable) "null token" with probability p during training. The model's parameters were updated on the MSE loss between predicted and true velocity.}
    \label{fig:Conditional_Training_loop}
\end{figure}

Figure~\ref{fig:Conditional_Training_loop} illustrates the conditional diffusion training pipeline. The process begins with raw input images (top left), which undergo standard data augmentations including random flipping, color jittering, and adjustments to sharpness and contrast. These augmentations increase visual diversity and improve model generalization. For each augmented image, a corresponding corrupted version is generated using the synthetic corruption function that simulates underwater degradation effects such as blurring, color distortion, and particulate occlusion.

Both augmented and corrupted images are then encoded into latent representations using the pretrained VAE. The diffusion model is trained to reconstruct the clean latent from its noisy counterpart, conditioned on the latent representation of the corrupted image. Conditioning is implemented by concatenating the two latent tensors and feeding the combined representation into the U-Net. The model learns to reverse the diffusion process by predicting the noise added to the clean latent at each timestep, with training guided by a mean squared error (MSE) loss between the predicted and true velocity vector.

To enable classifier-free guidance, random condition dropping is applied during training. With a fixed probability $p = 0.25$, the conditioning latent is replaced by a learnable “null token,” allowing the model to occasionally ignore conditioning information. This design improves model robustness and enables flexible operation in either guided or unguided modes during inference.

\begin{figure*}[tb]
    \centering
    \includegraphics[width=\linewidth]{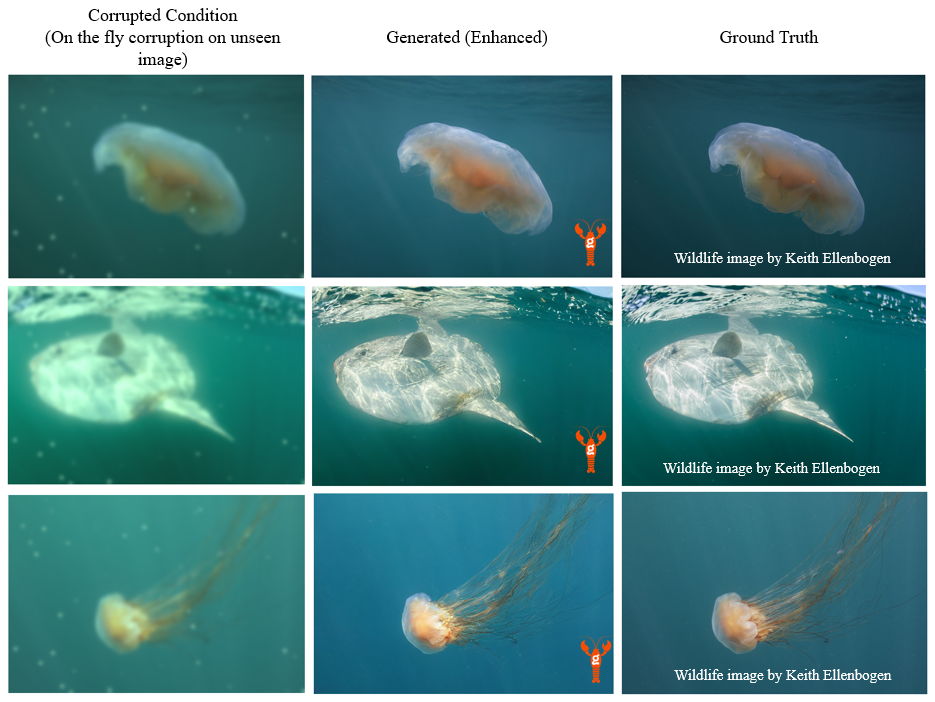}
    \caption{Examples of enhanced images from the held out evaluation set. Each row shows the input condition (left), the model's enhanced output (center), and the ground truth image (right).}
    \label{fig:Test_set_enhancement}
\end{figure*}

\subsection{Augmentations}

Training was done on $\sim$2500 moderately high-resolution images (512 x 768), consisting exclusively of Lion's Mane Jellyfish ($\sim$800), Blue Shark($\sim$1000), Mola Mola($\sim$400), and American Lobster ($\sim$300) images. The data were split 90/10 for training and evaluation, reserving $\sim$200 images for evaluation. An additional $\sim$50 images by Keith Ellenbogen consisting of species different from those the model saw in training were also used for testing.

A specific challenge to image-to-image latent diffusion is that artificial corruptions must be applied in the pixel space. For clarity, we underscore that corrupting the latent representations directly does not produce the effects of corrupting in the pixel space. As a result, corruption must be applied to the raw image data prior to encoding, which introduces a trade-off between computational efficiency and training dataset diversity.

Two initial strategies were explored to generate corrupted-clean image pairs. In the first approach, image corruption was performed on-the-fly within the training loop. While this ensured that the model was continually exposed to diverse generated corrupted samples, the computational cost was prohibitively high. The real-time application of the corruption function significantly slowed down training, making it infeasible for large-scale experiments. In the second approach, a fixed set of corrupted-clean pairs was precomputed and stored before training. Although this method greatly improved efficiency, it introduced risks of overfitting. The static nature of the precomputed dataset limited the diversity of corruptions seen during training, leading to the rapid memorization of patterns and diminishing generalization. Furthermore, precomputing a sufficiently large dataset (e.g., more than 10,000 pairs) introduced substantial storage and data management overhead.

To balance these considerations, an intermediate strategy was adopted. For each clean training image, a single corrupted counterpart was generated at the beginning of training and stored in the dataloader. Then, every 10 epochs, $10\%$ of the training data ($\sim$200 images selected at random) were refreshed by reapplying the corruption function. This periodic update mechanism enabled the model to benefit from a continually evolving training dataset without incurring the high computational costs. At the same time, the strategy mitigated overfitting by periodically introducing new corruption patterns, thus preserving sufficient data diversity throughout training. This compromise proved effective in maintaining training efficiency while also supporting robust generalization performance.

\subsection{Inference}

Once trained to reverse the synthetic corruption function, the conditional latent diffusion model could be guided in multiple ways during inference to enhance degraded underwater images. Given an underwater image, the conditioning latent could be constructed by encoding one of the following:

\begin{enumerate}
\item The raw underwater image.
\item The image after applying the synthetic corruption pipeline.
\item The image after applying a mild version of the synthetic corruption (e.g., reduced blur etc).
\end{enumerate}

Empirically, the choice of conditioning mode was found to influence enhancement quality depending on the distributional characteristics of the input. For in-distribution samples (images depicting marine species and visual conditions represented in the training set (e.g., sharks, jellyfish, mola mola, and lobster))Strategy 2 produced the best results. By reapplying the training corruption, the conditioning input remained within the learned distribution, enabling the model to effectively perform the denoising and restoration steps it was optimized for.

Conversely, for out-of-distribution samples involving unseen species or novel environmental conditions, Strategy 3 yielded superior performance. The mild corruption ensured that the conditioning latent remained near the training manifold without excessively distorting unfamiliar structures. This balance allowed the model to leverage its learned priors while preserving new semantic and spatial cues, avoiding over-correction or artifact formation.

\begin{figure}[tb]
    \centering
    \includegraphics[width=0.9\linewidth]{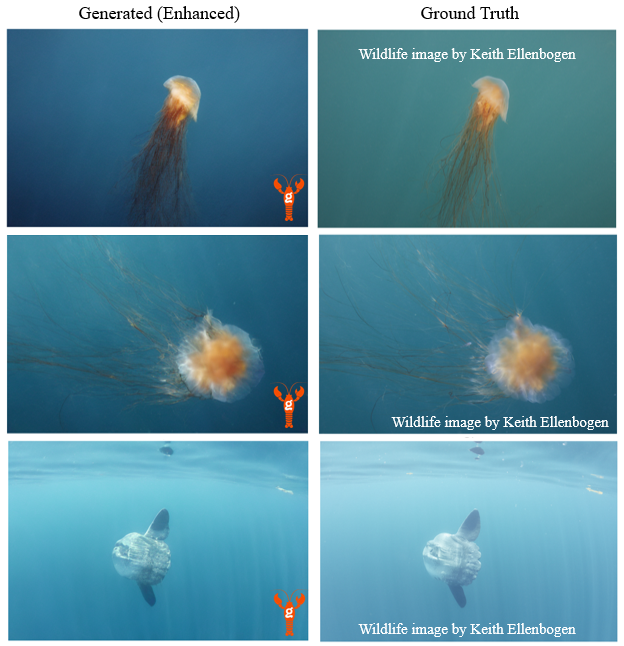}
    \caption{Example hue color correction on jellyfish and mola mola images.}
    \label{fig:Hue_correction}
\end{figure}

\section{Results}

In this section we present the capabilities and limitations of our trained diffusion model for image enhancement.

\subsection{In-distribution generalization}

In this section, we present image enhancement results using the generative model on samples from the held-out evaluation set of $\sim$ 200 images. These test samples depict species that were included during training and were captured under similar environmental conditions and artistic style. Applying the model can be though of as an "autoenhance" feature that can rapidly yield a unique enhanced version of a degraded image.

Figure \ref{fig:Test_set_enhancement} illustrates representative examples of the model’s enhancement capabilities. The top row features a jellyfish, where the image has been brightened and the animal’s coloration and contrast have been enhanced to make it stand out more distinctly. The middle row shows a mola mola, where the enhanced image closely resembles the ground truth but displays a notably different lighting pattern on the fish, demonstrating the model’s ability to render complex light patterns. In the bottom row, the image of a lion’s mane jellyfish highlight's the model's ability to correct underwater light scattering, generating a natural blue tone on the background water while improving contrast to better highlight the subject from its background.

 Figure \ref{fig:Hue_correction} showcases examples where the model successfully corrects pronounced hue shifts, caused by light absorption and scattering underwater. These corrections restore the natural coloration of marine subjects and their surroundings, improving visual fidelity and aesthetics. The enhanced images exhibit a more balanced color spectrum, which is particularly important for artistic presentation purposes. The top two rows include examples of Lion's Mane Jellyfish: the top image features a deep blue background with a nuanced gradient of blues replacing a flat green, upon which the jellyfish is gracefully overlaid. In the middle image, the gradient of blue tones is more pronounced, enhancing contrast and emphasizing the subject’s coloration so it stands out from the background. In the bottom image, the model reduces scattered white light throughout the scene, yielding a smoother and more balanced color spectrum that adds depth, visual substance, and detailed coloration to the composition.

\begin{figure}[b]
    \centering
    \includegraphics[width=\linewidth]{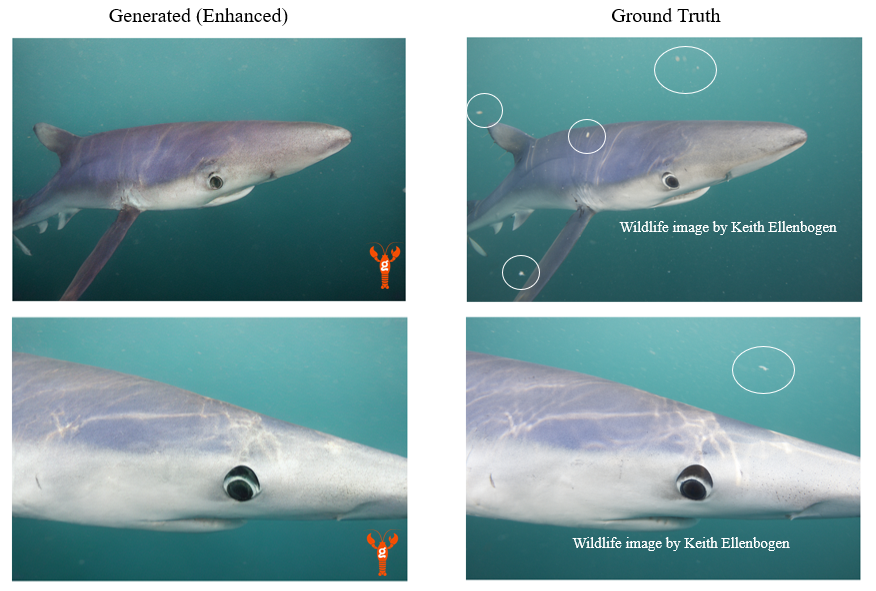}
    \caption{Example bubble correction on a blue shark images.}
    \label{fig:bubble_correction}
\end{figure}

\begin{figure}[tb]
    \centering
    \includegraphics[width=\linewidth]{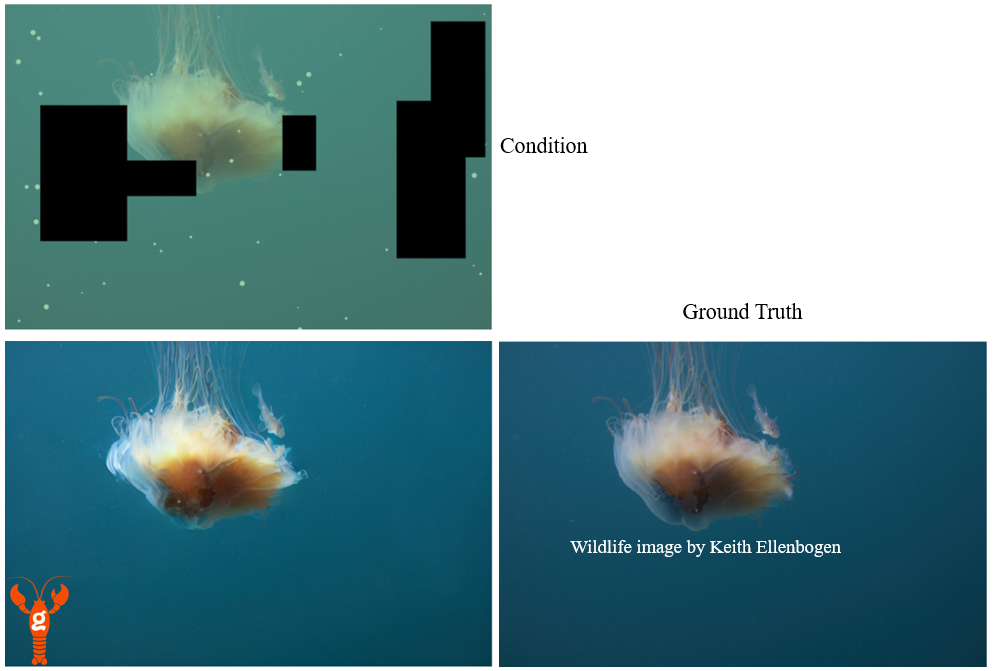}
    \caption{Inpainting a Jellyfish image. Condition is on the top left, generated image on the bottom left, and ground truth image on the bottom right.}
    \label{fig:Inpainting_Jelly_2}
\end{figure}

\begin{figure}[b]
    \centering
    \includegraphics[width=\linewidth]{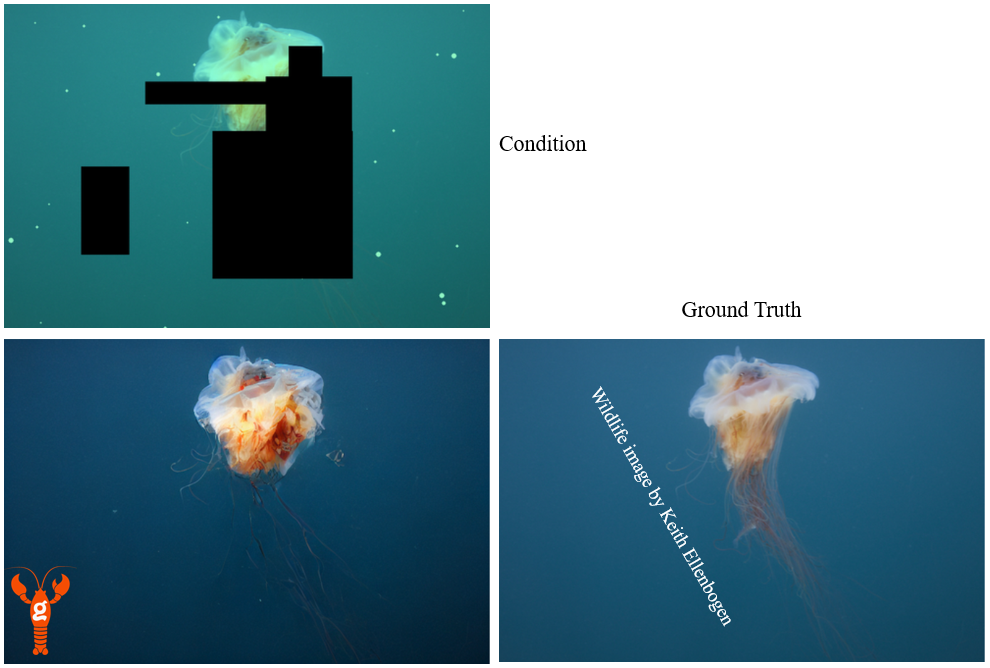}
    \caption{Inpainting a Jellyfish image. Condition is on the top left, generated image on the bottom left, and ground truth image on the bottom right.}
    \label{fig:eInpainting_Jellyfish_4}
\end{figure}

\begin{figure*}[tb]
    \centering
    \includegraphics[width=0.75\linewidth]{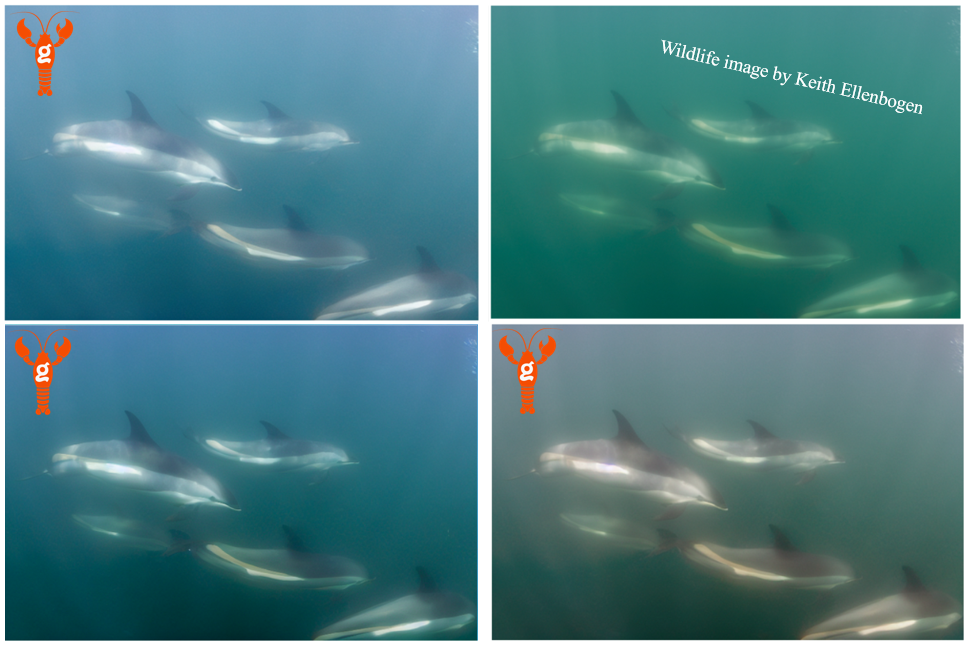}
    \caption{Color corrected image of four dolphins captured by Keith Ellenbogen. The model has not trained in dolphins or multiple subjects within the same image.}
    \label{fig:Dolphin_correction}
\end{figure*}

Figure~\ref{fig:bubble_correction} illustrates the model’s ability to suppress bubble artifacts commonly present in underwater photography. Both examples depict blue sharks captured under natural yet suboptimal conditions. In the top row, the model effectively removes bubbles overlapping the shark’s body and fins, restoring structural integrity and enhancing color contrast. In the bottom example, a large peripheral bubble is eliminated, yielding a cleaner and more visually coherent composition. These results demonstrate the model’s capacity to improve perceptual quality while preserving fine anatomical detail.

\begin{figure}[b]
    \centering
    \includegraphics[width=\linewidth]{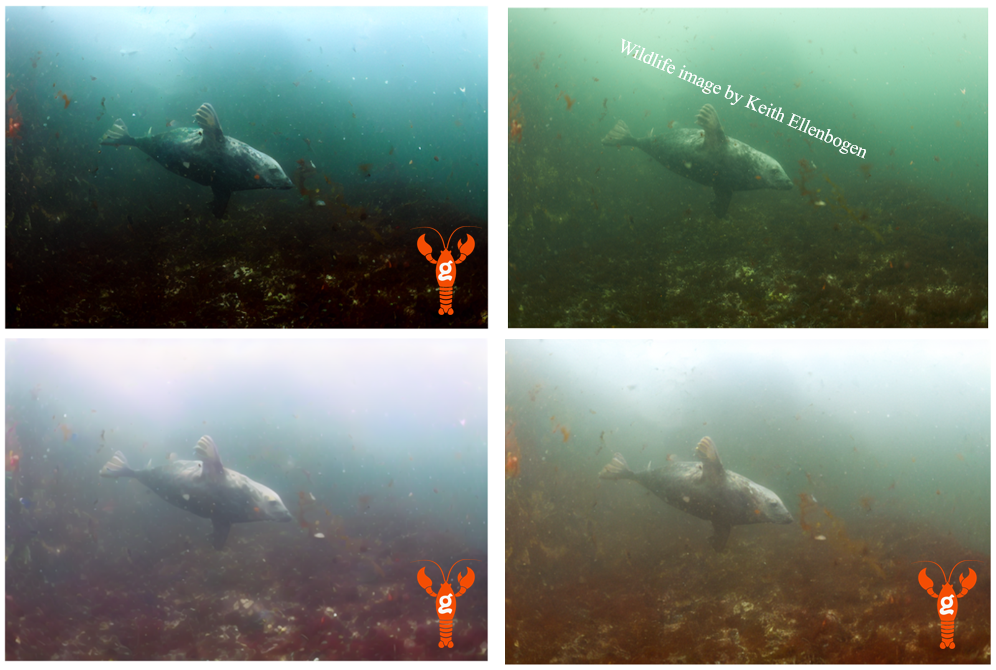}
    \caption{Example of generated enhancements of a seal image. Top right image is ground truth and the rest are generated.}
    \label{fig:four_seals}
\end{figure}

\begin{figure}[tb]
    \centering
    \includegraphics[width=\linewidth]{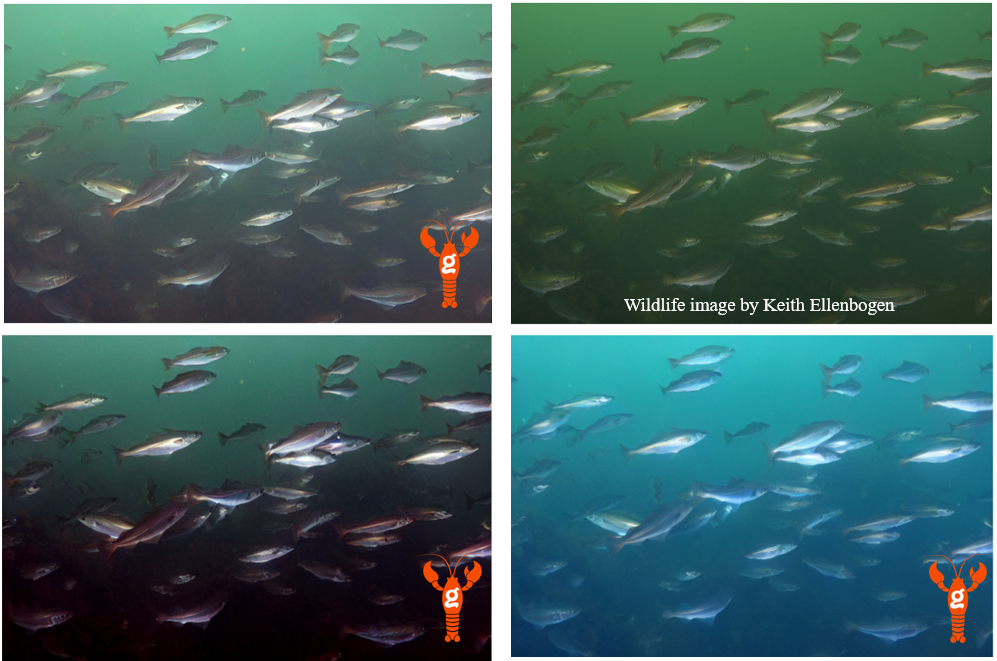}
    \caption{Example of generated enhancements of schooling fish image. Top right image is ground truth and the rest are generated.}
    \label{fig:four_fish_schools}
\end{figure}

\begin{figure}[b]
    \centering
    \includegraphics[width=\linewidth]{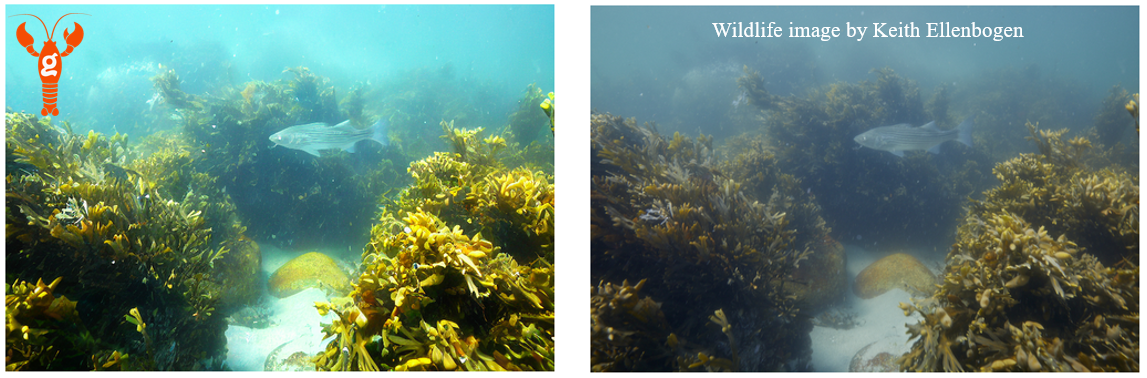}
    \caption{Example of generated enhancements of a complex scene. Right image is ground truth and the left is generated enhanced version.}
    \label{fig:bahamas_fish}
\end{figure}

\subsection{In-distribution inpainting}

In this section we present inpainting results of our trained models on Jellyfish. Figure \ref{fig:Inpainting_Jelly_2} illustrates a conditional inpainting example of on a Lion's Maine Jellyfish from the held out test set. The top-left panel shows the conditioning input, where large regions of the original image have been masked out. The bottom-left image shows the model’s reconstruction, and the right image is the ground truth reference. The model successfully restores the missing content with high structural and chromatic fidelity, accurately reproducing a jellyfish’s form and surrounding water tones. This demonstrates strong generative priors and effective conditional reasoning under substantial information loss.

Figure~\ref{fig:eInpainting_Jellyfish_4} presents an additional inpainting example on a jellyfish image from the test set. In this case, a substantial portion of the animal’s body was masked, requiring the model to reconstruct complex structural and textural details. The model successfully fills the missing region with anatomically plausible content, producing a coherent and visually consistent result. Notably, the generated image exhibits enhanced sharpness and color saturation, while synthetic bubble artifacts are effectively removed. These results further highlight the model’s ability to synthesize high-quality reconstructions under severe occlusion, demonstrating robustness and diversity in generative inpainting.

\begin{figure}[tb]
    \centering
    \includegraphics[width=\linewidth]{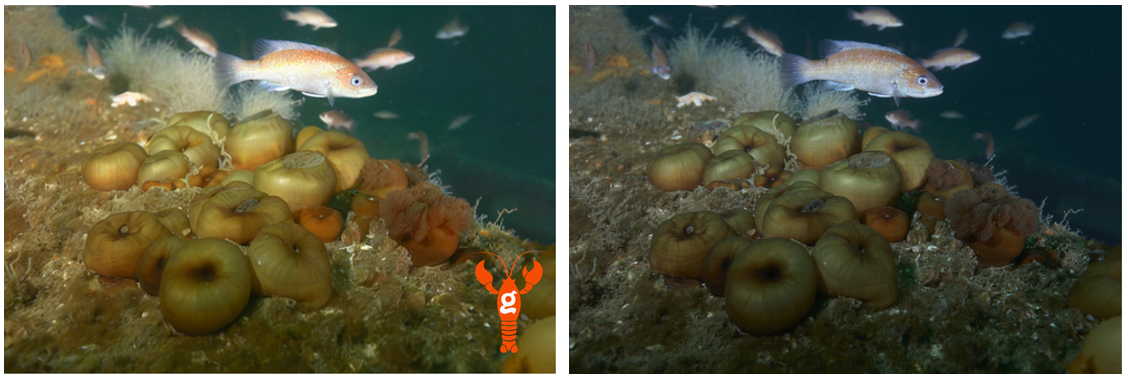}
    \caption{Example of generated enhancements of a complex scene. Right image is ground truth and the left is generated enhanced version.}
    \label{fig:scientific_enhancement}
\end{figure}

\begin{figure}[b]
    \centering
    \includegraphics[width=\linewidth]{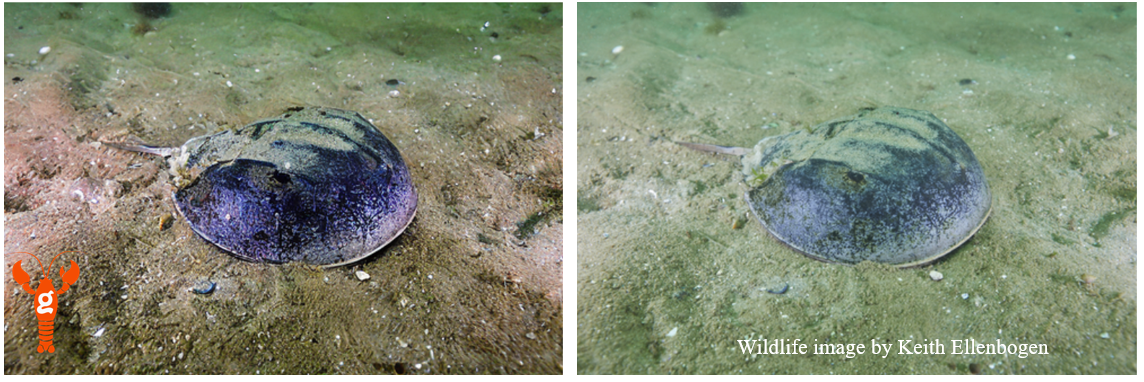}
    \caption{Example of generated enhancement on a horseshoe crab image. Right image is ground truth and the left is generated enhanced version.}
    \label{fig:horseshoe}
\end{figure}

\begin{figure}[tb]
    \centering
    \includegraphics[width=\linewidth]{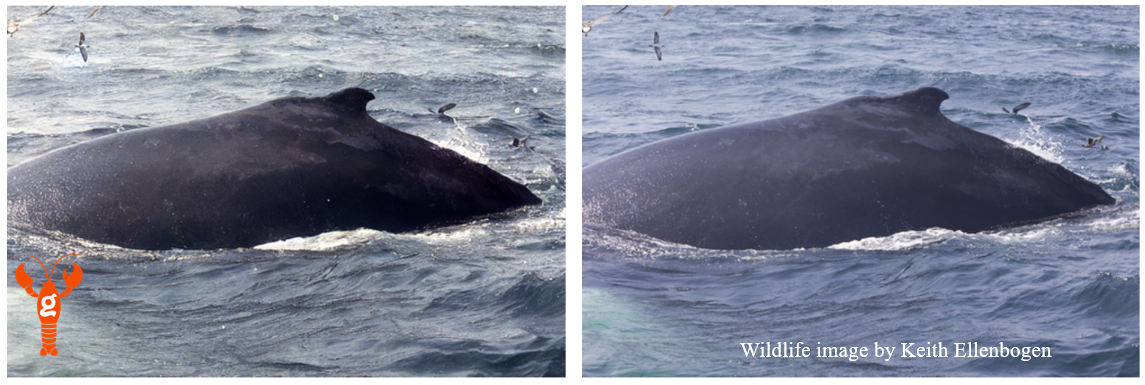}
    \caption{Example of generated enhancement on a whale image outside the water. Right image is ground truth and the left is generated enhanced version.}
    \label{fig:whale}
\end{figure}

\subsection{Out-of-distribution generalization}

In this section we present enhancement results on images by Keith Ellenbogen on species that were not seen by the model during training.

Figure~\ref{fig:Dolphin_correction} presents an enhancement example on a dolphin image captured within Massachusetts and the Gulf of Maine by Keith Ellenbogen. Although the model was never exposed to dolphins during training, it effectively generalizes to this unseen species, producing a visually coherent and natural enhancement.

The enhanced outputs (top left, bottom left, bottom right) are unique samples on the same condition exhibiting improved color balance relative to the original ground truth image (top right). The greenish cast characteristic of underwater imaging is corrected toward deeper blue tones, boosting aesthetics in ocean coloration and perceptual appeal. Beyond color restoration, the model enhances clarity and contrast—the outlines of the dolphins, particularly the leading individual, are sharper and more defined, increasing visual salience without introducing artifacts or hallucinations.

Importantly, the model preserves global lighting cues and compositional integrity. Subtle light gradients in the water column remain consistent, and the spatial arrangement and morphology of the animals are faithfully retained. These results demonstrate that the model not only restores degraded underwater scenes but does so in a manner consistent with the photographer’s original aesthetic and the physical properties of the environment. This example highlights the model’s strong generalization capability, improving image fidelity and visual quality even in fully out-of-distribution settings.

In this true inference sample the model's ability to enhance images for presentation purposes as well as for scientific purposes is evident.

Figures \ref{fig:four_seals} and \ref{fig:four_fish_schools} illustrate additional enhancement samples for seals and small fish schools. Both images are of species the model never encountered in training suffering from severe discoloration. The model successfully separates the animals from their background, pronounces their silhouettes by adequate recoloration, and adjusts the ocean background coloration restoring blue and red hues. Different outcomes on the same condition illustrate how the model can be used as an "autoenhance" feature for photographers who can rapidly generate numerous enhanced versions of images that would otherwise have to be manually post-processed.

Figures \ref{fig:bahamas_fish}-\ref{fig:whale} illustrate model performance on diverse underwater tasks on animals never encountered during training. In all figures the background is not clear ocean which introduces additional complexity in reproducing the complex underwater environments the model does not see during training. In all cases the model significantly improves the color spectrum distribution resulting in more visually appealing images that have increased detail and can be more easily used in presentations or for scientific exploration. 

\section{Conclusions}

We presented LOBSTgER-enhance, a conditional latent diffusion framework for underwater image restoration and enhancement. The model is trained on 512×768 images using a supervised dataset derived from a synthetic corruption process that emulates the characteristic degradations of underwater photography, including blurring, color distortion, and particulate occlusion. Despite being trained on a relatively small dataset of roughly 2.5k images spanning only four species, the model demonstrates strong generalization to diverse underwater conditions and to species never seen during training.

Empirical results show that LOBSTgER-enhance effectively restores color balance, sharpness, and structural integrity while maintaining both scientific realism and artistic intent. The model also exhibits robust inpainting capabilities, recovering missing regions and removing artifacts such as bubbles or particulate matter without introducing visual inconsistencies. These findings underscore the potential of diffusion-based conditional models to support high-fidelity image restoration in complex natural environments, while advancing computational tools for marine conservation and visual storytelling.

{
    \small
    \bibliographystyle{ieeenat_fullname}
    \bibliography{main}
}

\clearpage
\setcounter{page}{1}
\maketitlesupplementary
% -----------

\section{Linear vs. Cosine variance schedule}

Illustration of linear \cite{ho2020denoising} variance schedule and cosine \cite{nichol2021improved} variance schedule.

\begin{figure}[!htb]
    \centering
    \includegraphics[width=\linewidth]{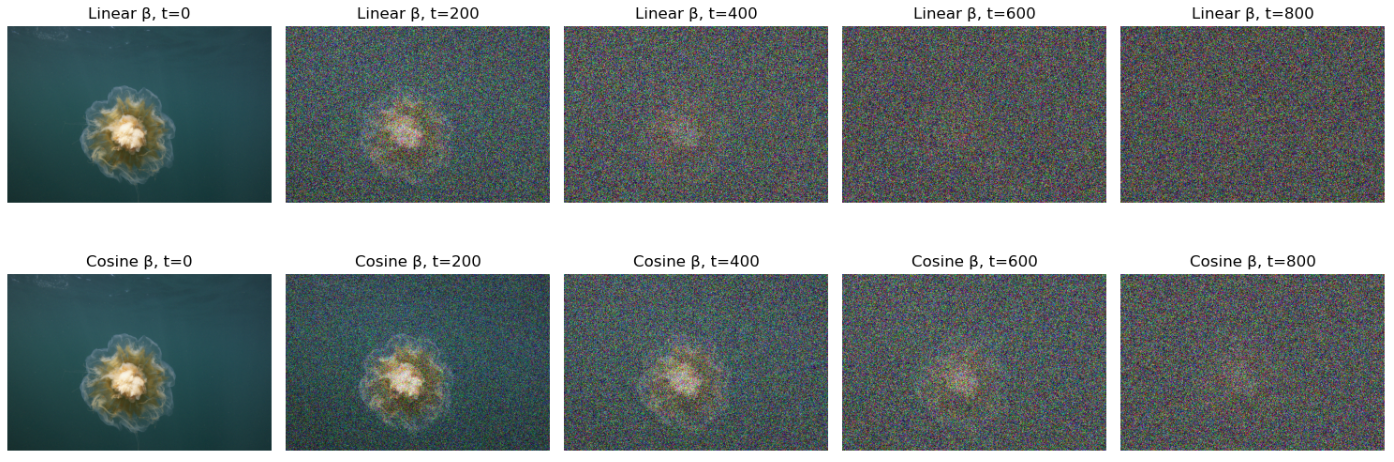}
    \caption{Noise addition for linear and cosine variance (noise) schedules at different time steps of the noise addition process. The linear schedule is shown in the top row and the cosine schedule on the bottom row. Evidently a cosine schedule degrades the image more progressively retaining more signal, especially at early timesteps.}
    \label{fig:Linear_v_cosine_schedule}
\end{figure}

\section{Additional Examples}
\subsection{In-sample-generalization}

In this section we present results from enhancing images from sepses seen in training but from the held out test set.

\begin{figure}[!htb]
    \centering
    \includegraphics[width=\linewidth]{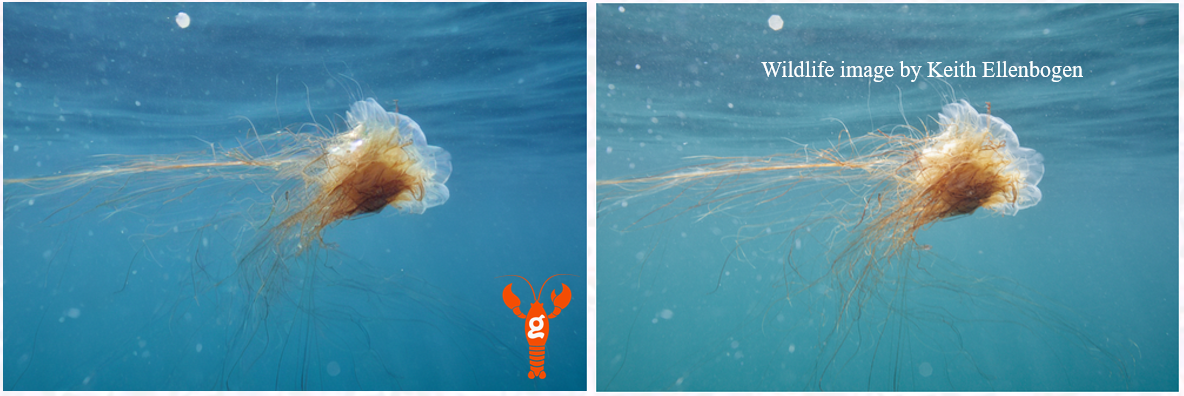}
    \caption{Example color enhanced jellyfish.}
    \label{fig:Color_correction_jellyfish}
\end{figure}

\begin{figure}[!htb]
    \centering
    \includegraphics[width=\linewidth]{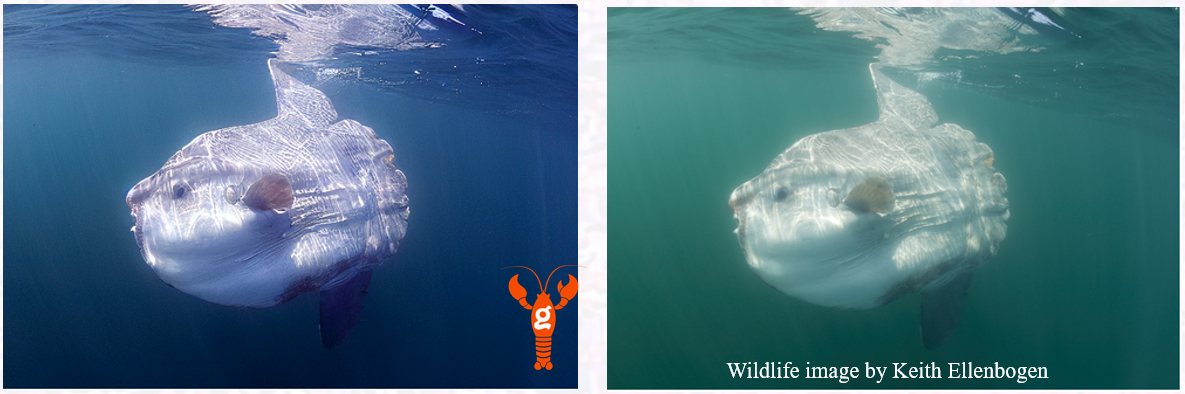}
    \caption{Example color enhanced mola mola.}
    \label{fig:enter-label}
\end{figure}

\begin{figure}[!htb]
    \centering
    \includegraphics[width=\linewidth]{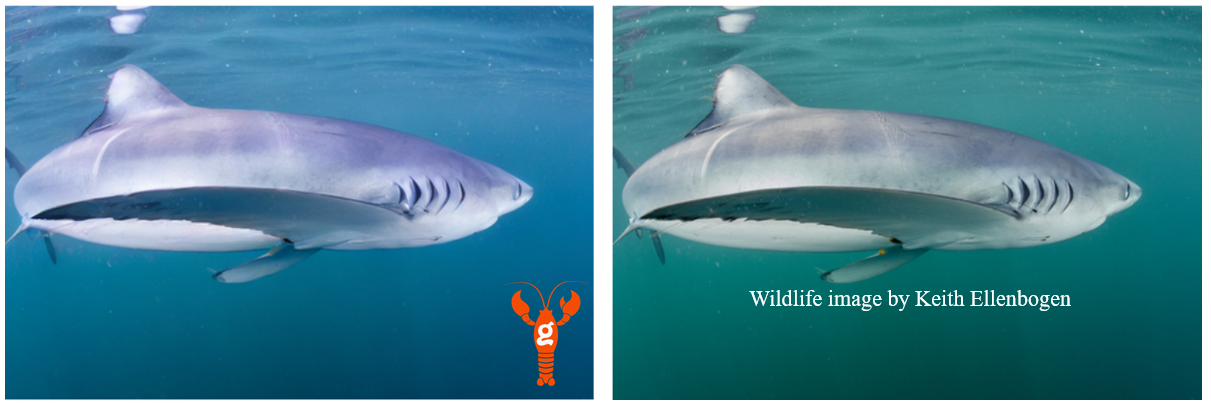}
    \caption{Example color enhanced blue shark.}
    \label{fig:enter-label}
\end{figure}

\begin{figure}[!htb]
    \centering
    \includegraphics[width=\linewidth]{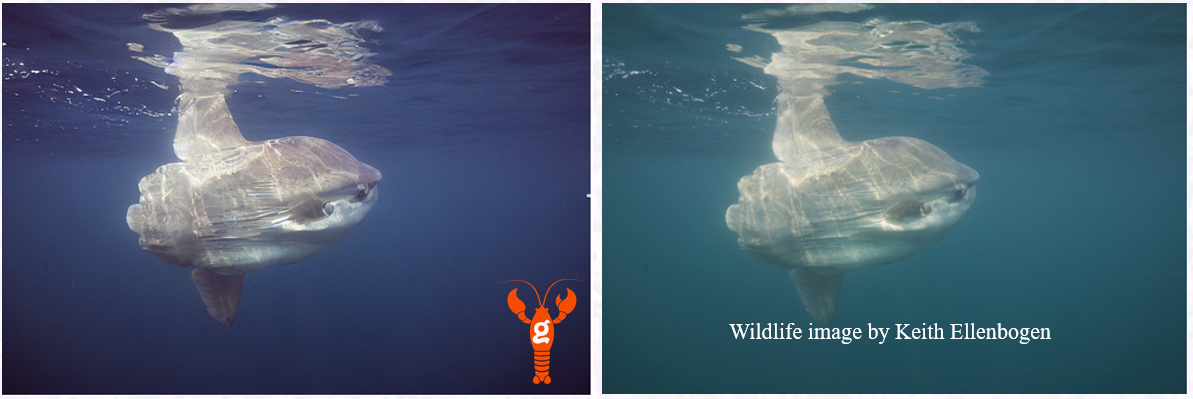}
    \caption{Example color enhanced mola mola.}
    \label{fig:enter-label}
\end{figure}

\begin{figure}[tb]
    \centering
    \includegraphics[width=\linewidth]{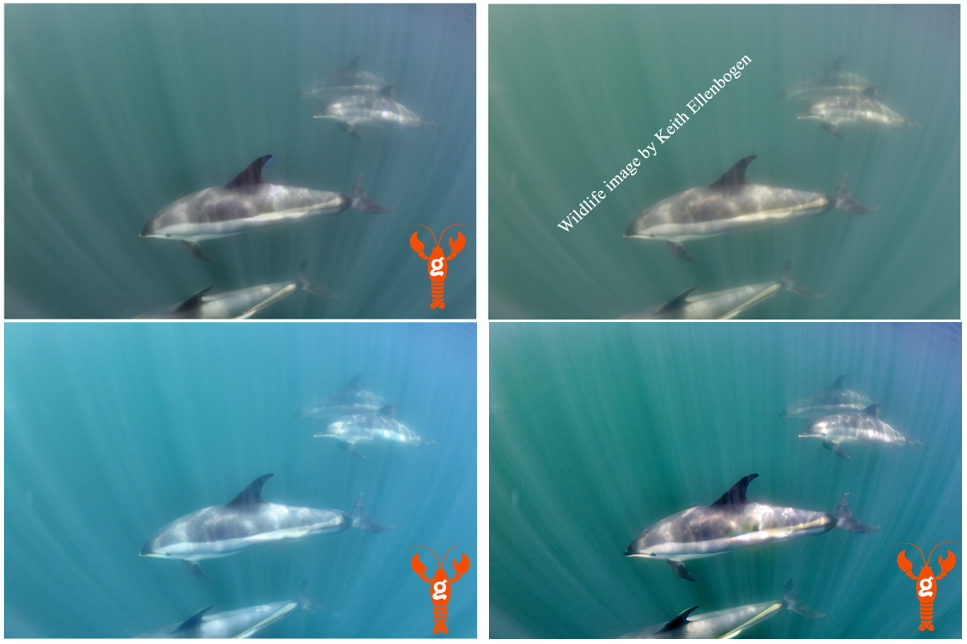}
    \caption{Example of generated enhancements of dolphins image. Top right image is ground truth and the rest are generated. Notice that the smaller model's outputs are more precise out of distribution compared to the bigger model.}
    \label{fig:Four_dolphins}
\end{figure}

\clearpage\newpage
\subsection{Inpainting samples}

\begin{figure}[!htb]
    \centering
    \includegraphics[width=\linewidth]{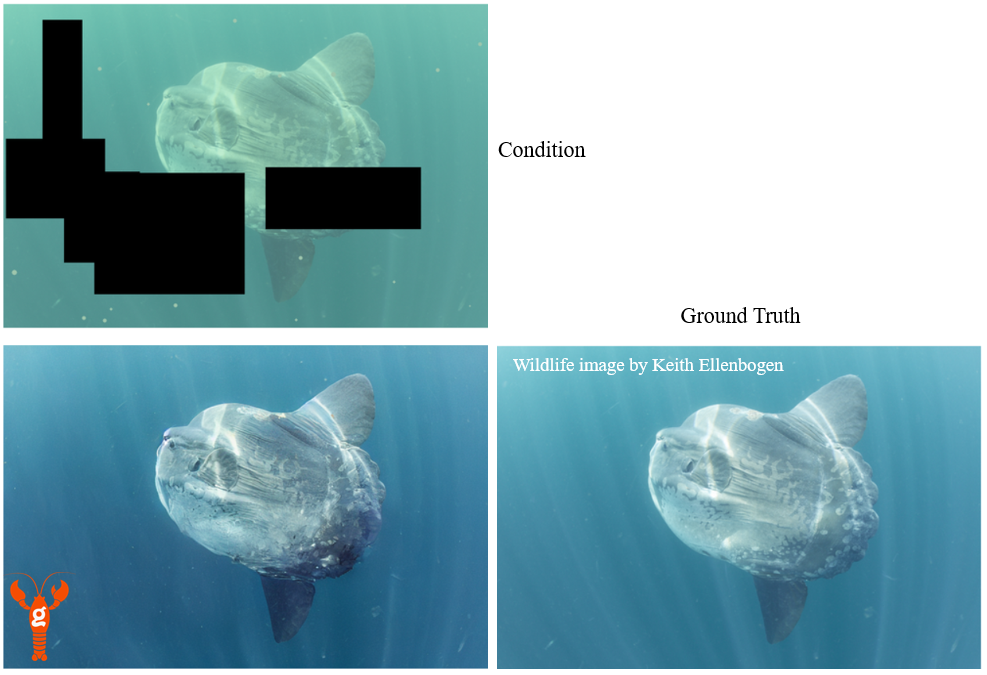}
    \caption{Inpainting a Mola Mola image. Condition is on the top left, generated image on the bottom left, and ground truth image on the bottom right.}
    \label{fig:enter-label}
\end{figure}

\begin{figure}[!htb]
    \centering
    \includegraphics[width=\linewidth]{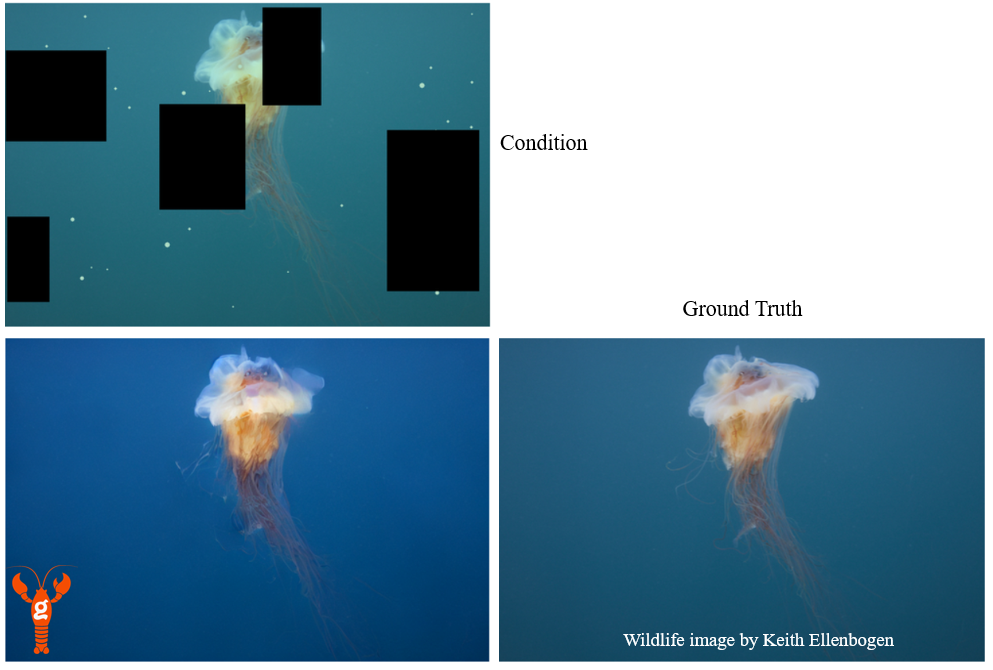}
    \caption{Inpainting a Jellyfish image. Condition is on the top left, generated image on the bottom left, and ground truth image on the bottom right.}
    \label{fig:enter-label}
\end{figure}

\begin{figure}[!htb]
    \centering
    \includegraphics[width=\linewidth]{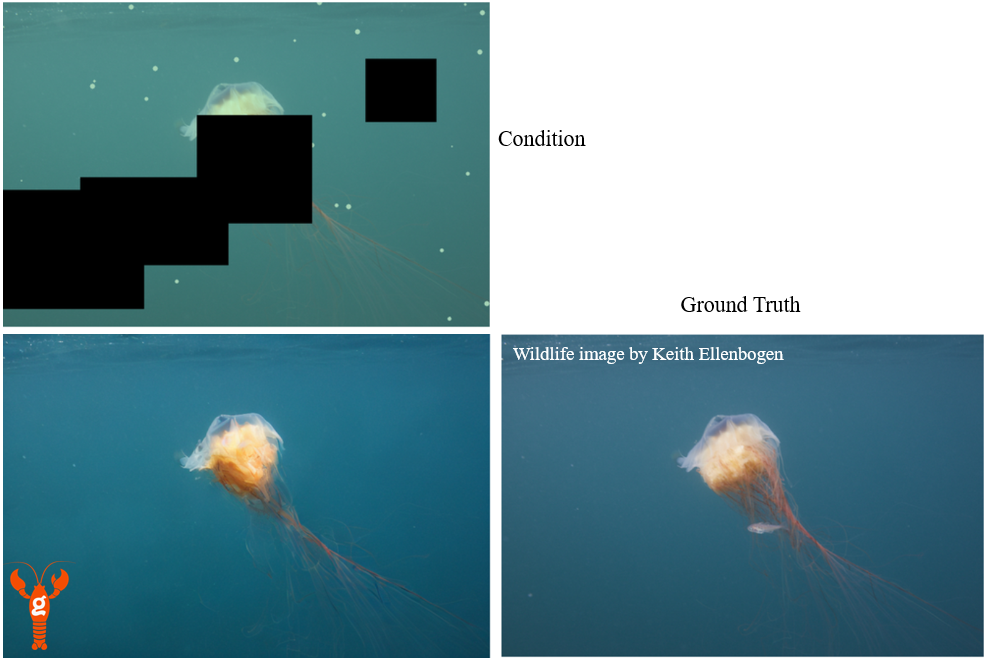}
    \caption{Inpainting a Jellyfish image. Condition is on the top left, generated image on the bottom left, and ground truth image on the bottom right.}
    \label{fig:enter-label}
\end{figure}

\begin{figure}[!htb]
    \centering
    \includegraphics[width=\linewidth]{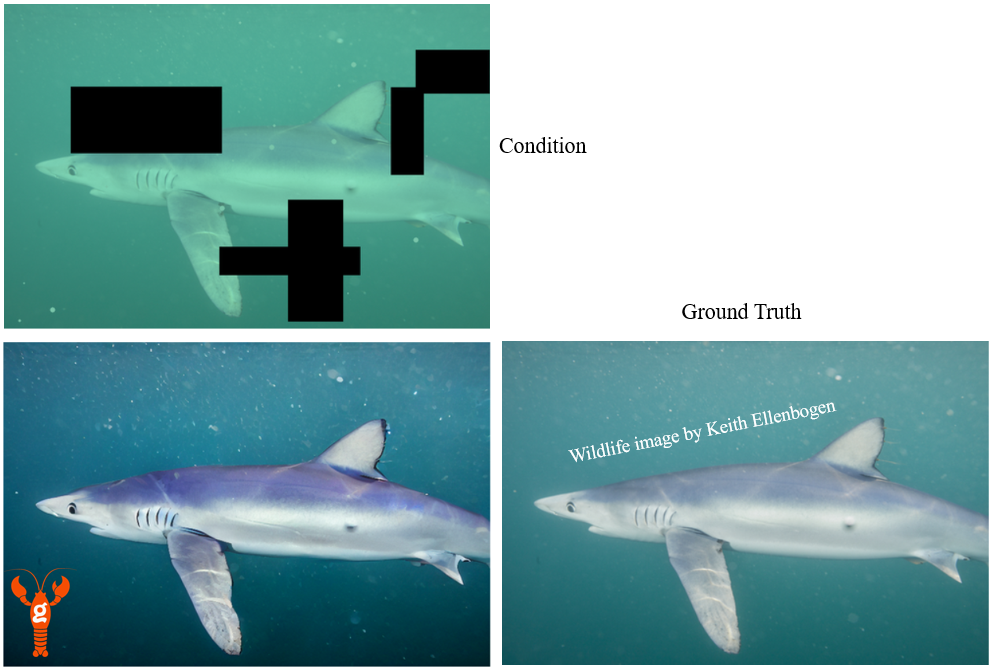}
    \caption{Inpainting a shark image. Condition is on the top left, generated image on the bottom left, and ground truth image on the bottom right.}
    \label{fig:enter-label}
\end{figure}

\begin{figure}[!htb]
    \centering
    \includegraphics[width=\linewidth]{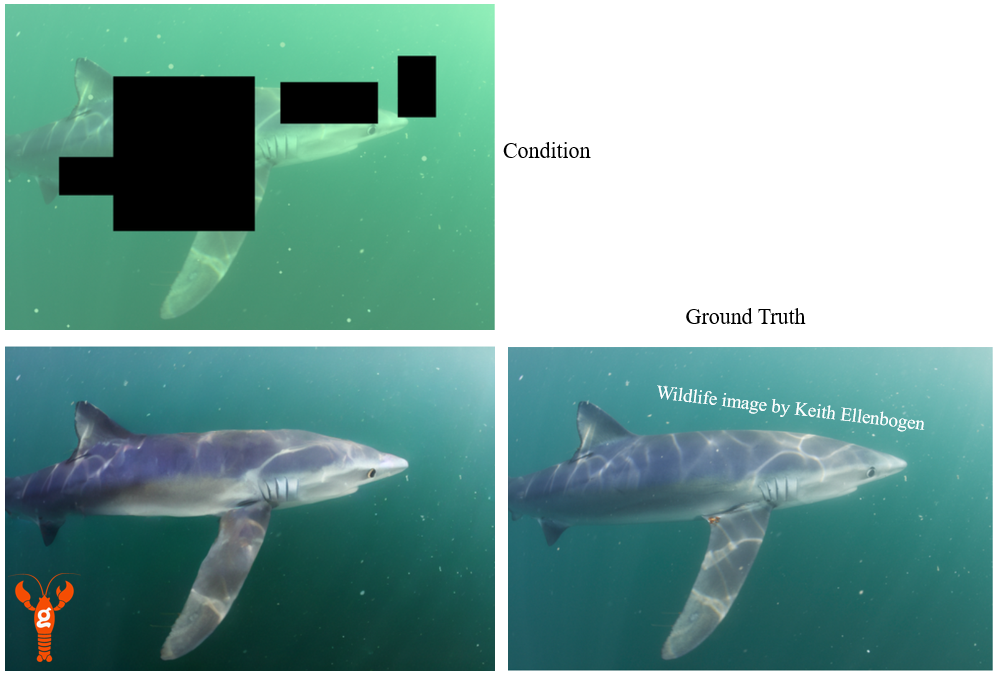}
    \caption{Inpainting a shark image. Condition is on the top left, generated image on the bottom left, and ground truth image on the bottom right.}
    \label{fig:enter-label}
\end{figure}

%%% -------------

\end{document}